\def\year{2018}\relax
\documentclass[letterpaper]{article} 
\usepackage{aaai18}  
\usepackage{times}  
\usepackage{helvet}  
\usepackage{courier}  
\usepackage{url}  
\usepackage{graphicx}  
\usepackage{amsmath,amssymb}
\usepackage{booktabs}
\usepackage{bm}

\usepackage[dvipsnames]{xcolor}

\def\S{\mathcal{S}}
\def\A{\mathcal{A}}
\def\R{r}
\def\z{\bm z}
\def\y{\bm y}
\def\b{\bm b}
\def\x{\bm x}
\def\W{\textbf{W}}
\def\p{\bm p}

\DeclareMathOperator*{\argmax}{\arg\!\max}

\begin{document}
%

\title{
Rainbow: Combining Improvements in Deep Reinforcement Learning
}
\author{Matteo Hessel\\ DeepMind \And Joseph Modayil\\ DeepMind \And Hado van Hasselt\\ DeepMind
\And Tom Schaul\\ DeepMind \And Georg Ostrovski\\ DeepMind \AND Will Dabney\\ DeepMind \And Dan Horgan\\ DeepMind \And Bilal Piot \\ DeepMind \And Mohammad Azar \\ DeepMind \And David Silver\\ DeepMind
}
\maketitle
\begin{abstract}
The deep reinforcement learning community has made several independent improvements to the DQN algorithm. However, it is unclear which of these extensions are complementary and can be fruitfully combined. This paper examines six extensions to the DQN algorithm and empirically studies their combination.  Our experiments show that the combination provides state-of-the-art performance on the Atari 2600 benchmark, both in terms of data efficiency and final performance. We also provide results from a detailed ablation study that shows the contribution of each component to overall performance.
\end{abstract}

\section{Introduction}

The many recent successes in scaling reinforcement learning (RL) to complex sequential decision-making problems were kick-started by the Deep Q-Networks algorithm (DQN; \citeauthor{Mnih2015} \citeyear{dqn-arxiv}, \citeyear{Mnih2015}). 
Its combination of Q-learning with convolutional neural networks and experience replay enabled it to learn, from raw pixels, how to play many Atari games at human-level performance.
Since then, many extensions have been proposed that enhance its speed or stability. 

Double DQN (DDQN; \citeauthor{van2016deep} \citeyear{van2016deep}) addresses an overestimation bias of Q-learning \cite{Hasselt2010double}, by decoupling selection and evaluation of the bootstrap action. Prioritized experience replay \cite{schaul2015prioritized} improves data efficiency, by replaying more often transitions from which there is more to learn. 
The dueling network architecture \cite{wang2016dueling} helps to generalize across actions by separately representing state values and action advantages. 
Learning from multi-step bootstrap targets \cite{Sutton:1988,Sutton:1998book}, as used in A3C \cite{Mnih:2016}, shifts the bias-variance trade-off and helps to propagate newly observed rewards faster to earlier visited states. 
Distributional Q-learning \cite{Bellemare2017ADP} learns a categorical distribution of discounted returns, instead of estimating the mean. 
Noisy DQN \cite{FortunatoAPMOGM17} uses stochastic network layers for exploration.
This list is, of course, far from exhaustive.

\begin{figure}[t!]
\centering
\vspace*{-1em}
\includegraphics[width=\columnwidth, trim={0 0 1cm 0},clip] {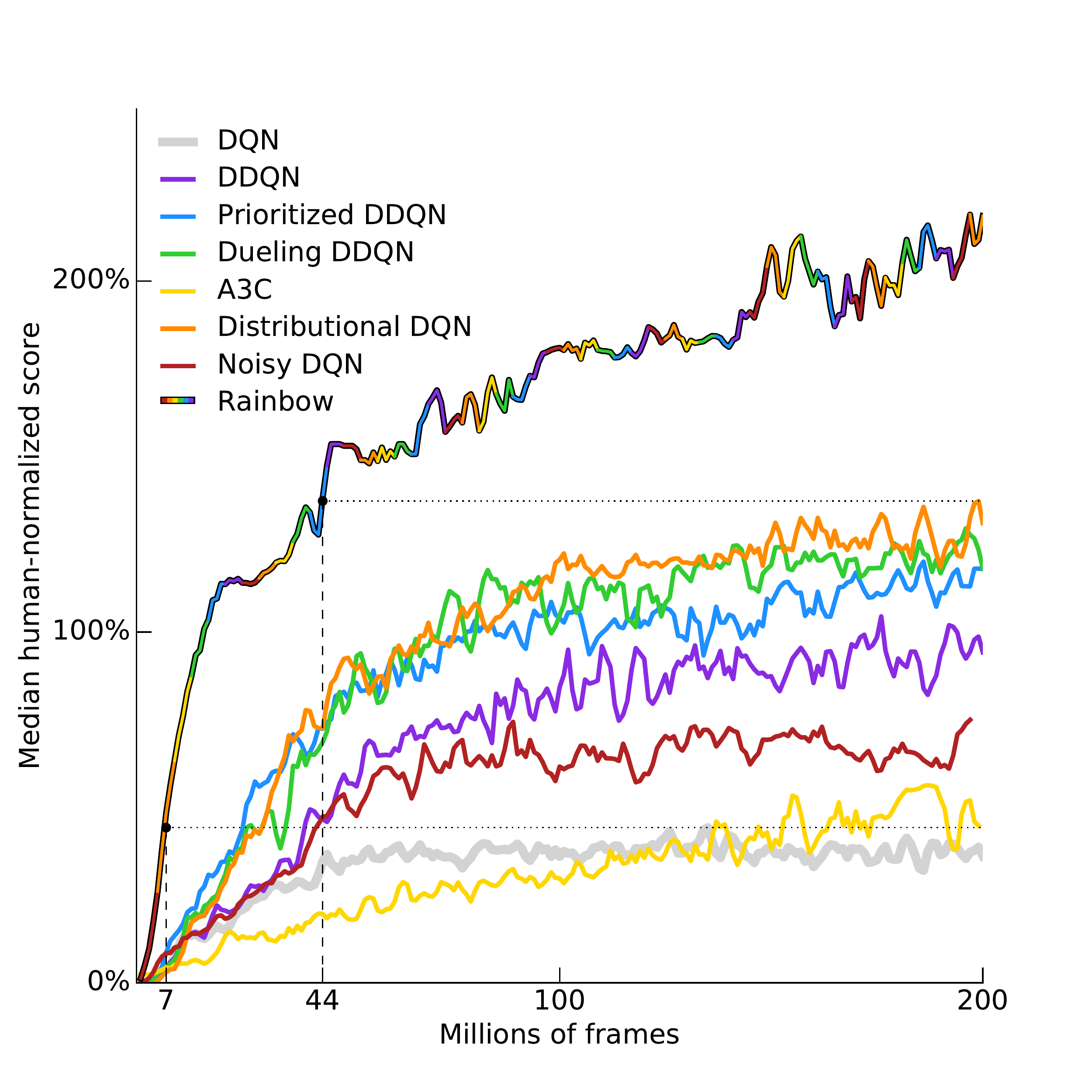}
\vspace{-2.5em}
\caption{\textbf{Median human-normalized performance} across 57 Atari games. We compare our integrated agent (rainbow-colored) to DQN (grey) and six published baselines. Note that we match DQN's best performance after 7M frames, surpass any baseline within 44M frames, and reach substantially improved final performance. Curves are smoothed with a moving average over 5 points.}
\label{fig:frontpage}
\end{figure}

Each of these algorithms enables substantial performance improvements in isolation. Since they do so by addressing radically different issues, and since they build on a shared framework, they could plausibly be combined.
In some cases this has been done: Prioritized DDQN and Dueling DDQN both use double Q-learning, and Dueling DDQN was also combined with prioritized experience replay.
In this paper we propose to study an agent that combines all the aforementioned ingredients. We show how these different ideas can be integrated, and that they are indeed largely complementary. In fact, their combination results in new state-of-the-art results on the benchmark suite of 57 Atari 2600 games from the Arcade Learning Environment \cite{bellemare2013arcade}, both in terms of data efficiency and of final performance. Finally we show results from ablation studies to help understand the contributions of the different components.

\section{Background}
Reinforcement learning addresses the problem of an \textit{agent} learning to act in an \textit{environment} in order to maximize a scalar \textit{reward} signal. No direct supervision is provided to the agent, for instance it is never directly told the best action. 

\paragraph{Agents and environments.}
At each discrete time step $t=0, 1, 2\ldots$, the environment provides the agent with an observation $S_t$, the agent responds by selecting an action $A_t$, and then the environment provides the next reward $R_{t+1}$, discount $\gamma_{t+1}$, and state $S_{t+1}$.
This interaction is formalized as a {\em Markov Decision Process}, or MDP,  which is a tuple $\langle \S, \A, T, \R, \gamma \rangle$, where $\S$ is a finite set of states, $\A$ is a finite set of actions, $T(s, a, s') = P[S_{t+1}=s'\mid S_t=s,A_t=a]$ is the (stochastic) transition function, $\R(s,a) = \mathbb{E}[R_{t+1} \mid S_t=s, A_t=a]$ is the reward function, and $\gamma \in [0,1]$ is a discount factor. In our experiments MDPs will be \textit{episodic} with a constant $\gamma_t = \gamma$, except on episode termination where $\gamma_t=0$, but the algorithms are expressed in the general form.

On the agent side, action selection is given by a policy $\pi$ that defines a probability distribution over actions for each state. From the state $S_t$ encountered at time $t$, we define the discounted return $G_t =\sum_{k=0}^{\infty} {\gamma_t^{(k)} R_{t+k+1}}$ as the discounted sum of future rewards collected by the agent, where the discount for a reward $k$ steps in the future is given by the product of discounts before that time, $\gamma_t^{(k)} = \prod_{i=1}^{k} \gamma_{t+i}$. An agent aims to maximize the expected discounted return by finding a good policy. 

The policy may be learned directly, or it may be constructed as a function of some other learned quantities. In value-based reinforcement learning, the agent learns an estimate of the expected discounted return, or value, when following a policy $\pi$ starting from a given state, $v^\pi(s)=E_\pi[G_t| S_t =s]$, or state-action pair, $q^\pi(s,a) = E_\pi[G_t|S_t=s, A_t =a]$. A common way of deriving a new policy from a state-action value function is to act $\epsilon$-greedily with respect to the action values. This corresponds to taking the action with the highest value (the \textit{greedy} action) with probability $(1-\epsilon)$, and to otherwise act uniformly at random with probability $\epsilon$. Policies of this kind are used to introduce a form of \textit{exploration}: by randomly selecting actions that are sub-optimal according to its current estimates, the agent can discover and correct its estimates when appropriate. The main limitation is that it is difficult to discover alternative courses of action that extend far into the future; this has motivated research on more directed forms of exploration.

\paragraph{Deep reinforcement learning and DQN.}
Large state and/or action spaces make it intractable to learn Q value estimates for each state and action pair independently. 
In deep reinforcement learning, we represent the various components of agents, such as policies $\pi(s,a)$ or values $q(s,a)$, with deep (i.e., multi-layer) neural networks. The parameters of these networks are trained by gradient descent to minimize some suitable loss function.

In DQN \cite{Mnih2015} deep networks and reinforcement learning were successfully combined by using a convolutional neural net to approximate the action values for a given state $S_t$ (which is fed as input to the network in the form of a stack of raw pixel frames). At each step, based on the current state, the agent selects an action $\epsilon$-greedily with respect to the action values, and adds a transition ($S_t, A_t, R_{t+1}, \gamma_{t+1}, S_{t+1}$) to a replay memory buffer \cite{Lin1992}, that holds the last million transitions.  The parameters of the neural network are optimized by using stochastic gradient descent to minimize the loss
\begin{equation} \label{Qlearning}
(R_{t+1} + \gamma_{t+1} \max_{a'} q_{\overline{\theta}}(S_{t+1}, a') - q_\theta(S_t, A_t))^2 \ ,
\end{equation}
where $t$ is a time step randomly picked from the replay memory.
The gradient of the loss is back-propagated only into the parameters $\theta$ of the \textit{online network} (which is also used to select actions); the term $\overline{\theta}$ represents the parameters of a \textit{target network}; a periodic copy of the online network which is not directly optimized. The optimization is performed using RMSprop \cite{tieleman2012lecture}, a variant of stochastic gradient descent, on mini-batches sampled uniformly from the experience replay. This means that in the loss above, the time index $t$ will be a random time index from the last million transitions, rather than the current time. The use of experience replay and target networks enables relatively stable learning of Q values, and led to super-human performance on several Atari games.

\section{Extensions to DQN}
DQN has been an important milestone, but several limitations of this algorithm are now known, and many extensions have been proposed.
We propose a selection of six extensions that each have addressed a limitation and improved overall performance. 
To keep the size of the selection manageable, we picked a set of extensions that address distinct concerns (e.g., just one of the many addressing exploration).

\paragraph{Double Q-learning.}
Conventional Q-learning is affected by an overestimation bias, due to the maximization step in Equation \ref{Qlearning}, and this can harm learning. Double Q-learning \cite{Hasselt2010double}, addresses this overestimation by decoupling, in the maximization performed for the bootstrap target, the selection of the action from its evaluation. It is possible to effectively combine this with DQN \cite{van2016deep}, using the loss
\[  
(R_{t+1} + \gamma_{t+1} q_{\overline{\theta}}(S_{t+1}, \argmax_{a'}q_\theta(S_{t+1}, a')) - q_\theta(S_t, A_t))^2.
\]  
This change was shown to reduce harmful overestimations that were present for DQN, thereby improving performance.

\paragraph{Prioritized replay.}
DQN samples uniformly from the replay buffer. Ideally, we want to sample more frequently those transitions from which there is much to learn. As a proxy for learning potential, prioritized experience replay \cite{schaul2015prioritized} samples transitions with probability $p_t$ relative to the last encountered absolute \textit{TD error}:
\[  
p_t \propto  \left|R_{t+1} + \gamma_{t+1} \max_{a'} q_{\overline{\theta}}(S_{t+1}, a') - q_\theta(S_t, A_t)\right|^{\omega} \,,
\]  
where $\omega$ is a hyper-parameter that determines the shape of the distribution.
New transitions are inserted into the replay buffer with maximum priority, providing a bias towards recent transitions. Note that stochastic transitions might also be favoured, even when there is little left to learn about them.

\paragraph{Dueling networks.}
The dueling network is a neural network architecture designed for value based RL. It features two streams of computation, the value and advantage streams, sharing a convolutional encoder, and merged by a special aggregator \cite{wang2016dueling}. This corresponds to the following factorization of action values:
\[  
q_\theta(s,a) = v_{\eta}(f_\xi(s)) + a_{\psi}(f_\xi(s),a) - \frac{\sum_{a'}a_{\psi}(f_\xi(s),a')}{N_\text{actions}},
\]  
where $\xi$, $\eta$, and $\psi$ are, respectively, the parameters of the shared encoder $f_\xi$, of the value stream $v_\eta$, and of the advantage stream $a_\psi$; and $\theta = \{\xi, \eta, \psi\}$ is their concatenation.

\paragraph{Multi-step learning.}
Q-learning accumulates a single reward and then uses the greedy action at the next step to bootstrap. Alternatively, forward-view \textit{multi-step} targets can be used~\cite{Sutton:1988}.  We define the truncated $n$-step return from a given state $S_t$ as
\begin{equation}\label{multiR}
R_t^{(n)} \equiv \sum_{k=0}^{n-1} {\gamma_t^{(k)} R_{t+k+1}} \,.
\end{equation}
A multi-step variant of DQN is then defined by minimizing the alternative loss, 
\[  
(R_t^{(n)} +\gamma_t^{(n)}  \max_{a'} q_{\overline{\theta}}(S_{t+n},a') - q_\theta(S_t, A_t))^2.
\]  
\noindent Multi-step targets with suitably tuned $n$ often lead to faster learning~\cite{Sutton:1998book}.

\paragraph{Distributional RL.}
We can learn to approximate the distribution of returns instead of the expected return.  Recently Bellemare, Dabney, and Munos (2017)\nocite{Bellemare2017ADP} proposed to model such distributions with probability masses placed on a discrete support ${\bm z}$, where $\z$ is a vector with $N_{\text{atoms}} \in \mathbb{N}^+$ \emph{atoms}, defined by $z^i = v_{\min} + (i-1)\frac{v_{\max}-v_{\min}}{N_{\text{atoms}}-1}$ for $i \in \{1, \ldots, N_{\text{atoms}}\}$.  The approximating distribution $d_t$ at time $t$ is defined on this support, with the probability mass $p^i_{\theta}(S_t, A_t)$ on each atom $i$, such that $d_t = (\z, \p_{\theta}(S_t, A_t))$.  The goal is to update $\theta$ such that this distribution closely matches the actual distribution of returns.

To learn the probability masses, the key insight is that return distributions satisfy a variant of Bellman's equation. For a given state $S_{t}$ and action $A_{t}$, the distribution of the returns under the optimal policy $\pi^*$ should match a target distribution defined by taking the distribution for the next state $S_{t+1}$ and action $a^*_{t+1} = \pi^*(S_{t+1})$, contracting it towards zero according to the discount, and shifting it by the reward (or distribution of rewards, in the stochastic case).  A distributional variant of Q-learning is then derived by first constructing a new support for the target distribution, and then minimizing the Kullbeck-Leibler divergence between the distribution $d_t$ and the target distribution $d'_t \equiv (R_{t+1} + \gamma_{t+1} \z, ~~\p_{\overline{\theta}}(S_{t+1}, \overline{a}^*_{t+1}))$,
\begin{equation} \label{DistribLoss}
D_{\text{KL}}(\Phi_{\z} d'_t || d_t) \,.
\end{equation}
Here $\Phi_{\z}$ is a L2-projection of the target distribution onto the fixed support $\z$, and $\overline{a}^*_{t+1} = \argmax_a q_{\overline{\theta}}(S_{t+1}, a)$ is the greedy action with respect to the mean action values $q_{\overline{\theta}}(S_{t+1}, a) = \z^{\top} \p_{\theta}(S_{t+1}, a)$ in state $S_{t+1}$.

As in the non-distributional case, we can use a frozen copy of the parameters $\overline{\theta}$ to construct the target distribution. The parametrized distribution can be represented by a neural network, as in DQN, but with $N_{\text{atoms}} \times N_{\text{actions}}$ outputs. A \textit{softmax} is applied independently for each action dimension of the output to ensure that the distribution for each action is appropriately normalized.

\paragraph{Noisy Nets.}
The limitations of exploring using $\epsilon$-greedy policies are clear in games such as Montezuma's Revenge, where many actions must be executed to collect the first reward. Noisy Nets \cite{FortunatoAPMOGM17} propose a noisy linear layer that combines a deterministic and noisy stream,
\begin{equation} \label{Noisy}
\y = (\b + \W \x) + (\b_{noisy} \odot \epsilon^b + (\W_{noisy} \odot  \epsilon^w)\x),
\end{equation}
\noindent where $\epsilon^b$ and $\epsilon^w$ are random variables, and $\odot$ denotes the element-wise product. This transformation can then be used in place of the standard linear $\y = \b + \W \x$. Over time, the network can learn to ignore the noisy stream, but will do so at different rates in different parts of the state space, allowing state-conditional exploration with a form of self-annealing. 

\section{The Integrated Agent} \label{KS}

In this paper we integrate all the aforementioned components into a single integrated agent, which we call \textit{Rainbow}.

First, we replace the 1-step distributional loss \eqref{DistribLoss} with a multi-step variant. We construct the target distribution by contracting the value distribution in $S_{t+n}$ according to the cumulative discount, and shifting it by the truncated $n$-step discounted return. This corresponds to defining the target distribution as $d^{(n)}_t = (R^{(n)}_{t} + \gamma_t^{(n)} \z,~~ \p_{\overline{\theta}}(S_{t+n}, a^*_{t+n}))$. The resulting loss is
\[  
D_{\text{KL}}(\Phi_{\z} d_{t}^{(n)} || d_t) \,,
\]  
where, again, $\Phi_{\z}$ is the projection onto $\z$.

We combine the multi-step distributional loss with double Q-learning by
using the greedy action in $S_{t+n}$ selected according to the \emph{online network} as the bootstrap action $a^*_{t+n}$, and evaluating such action using the \emph{target network}.

In standard proportional prioritized replay \cite{schaul2015prioritized} the absolute TD error is used to prioritize the transitions. This can be computed in the distributional setting, using the mean action values. However, in our experiments all distributional Rainbow variants prioritize transitions by the KL loss, since this is what the algorithm is minimizing:
\[  
p_t \propto \left( D_{\text{KL}}(\Phi_{\z}d_{t}^{(n)} || d_t) \right)^{\omega} \,.
\]  
The KL loss as priority might be more robust to noisy stochastic environments because the loss can continue to decrease even when the returns are not deterministic.

The network architecture is a dueling network architecture adapted for use with return distributions. The network has a shared representation $f_{\xi}(s)$, which is then fed into a value stream $v_{\eta}$ with $N_{\text{atoms}}$ outputs, and into an advantage stream $a_{\xi}$ with $N_{\text{atoms}} \times N_\text{actions}$ outputs, where $a^i_{\xi}(f_{\xi}(s), a)$ will denote the output corresponding to atom $i$ and action $a$.  For each atom $z^i$, the value and advantage streams are aggregated, as in dueling DQN, and then passed through a softmax layer to obtain the normalised parametric distributions used to estimate the returns' distributions:
\[
p^i_{\theta}(s, a) = \frac{\exp(v^i_{\eta}(\phi) + a^i_{\psi}(\phi, a) - \overline{a}^i_{\psi}(s))}{\sum_j \exp(v^j_{\eta}(\phi) + a^j_{\psi}(\phi, a) - \overline{a}^j_{\psi}(s))} \,,
\]
where $\phi = f_{\xi}(s)$ and $\overline{a}^i_{\psi}(s) = \frac{1}{N_{\text{actions}}} \sum_{a'} a^i_{\psi}(\phi, a')$.

We then replace all linear layers with their noisy equivalent described in Equation \eqref{Noisy}. Within these noisy linear layers we use factorised Gaussian noise \cite{FortunatoAPMOGM17} to reduce the number of independent noise variables. 

\begin{figure*}
\centering
\includegraphics[width=\textwidth,trim={0 1.1cm 0.4cm 0},clip]{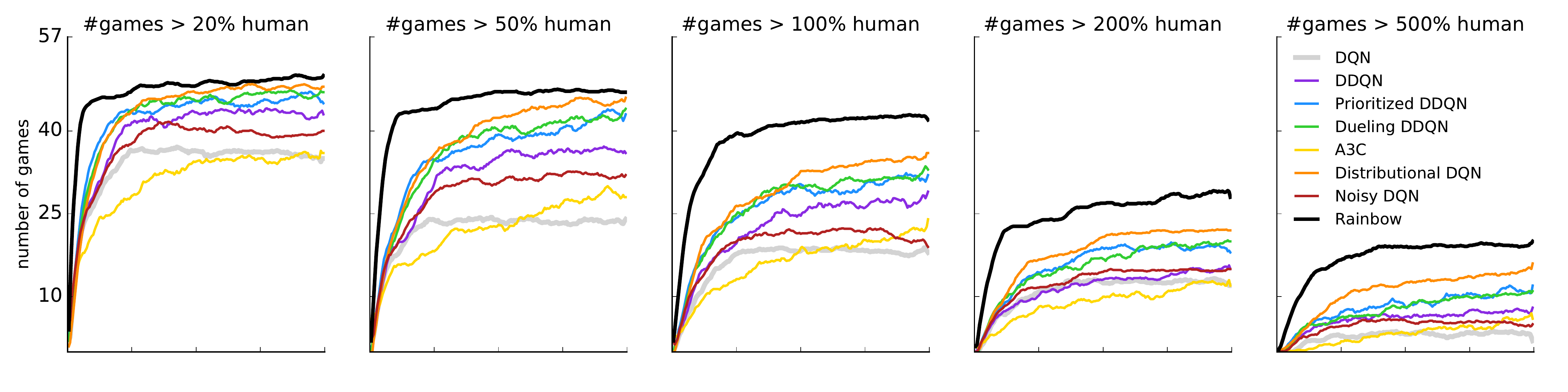}
\includegraphics[width=\textwidth]{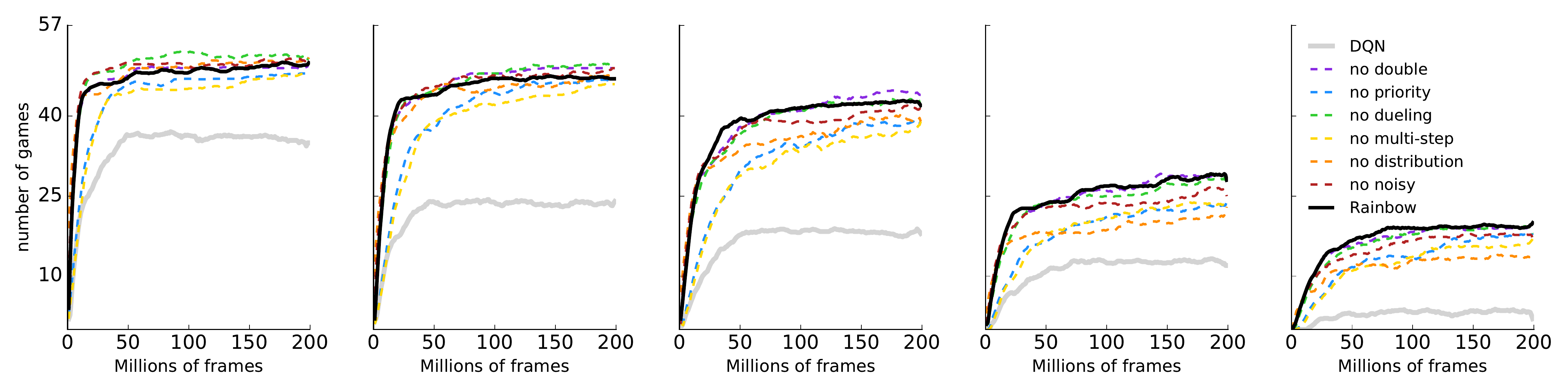}
\vspace*{-2em}
\caption{Each plot shows, for several agents, the number of games where they have achieved at least a given fraction of human performance, as a function of time. From left to right we consider the 20\%, 50\%, 100\%, 200\% and 500\% thresholds. On the first row we compare Rainbow to the baselines. On the second row we compare Rainbow to its ablations.
\label{fig:num_games}}
\end{figure*}

\section{Experimental Methods}

We now describe the methods and setup used for configuring and evaluating the learning agents.

\paragraph{Evaluation Methodology.}
We evaluated all agents on 57 Atari 2600 games from the arcade learning environment \cite{bellemare2013arcade}. We follow the training and evaluation procedures of \citeauthor{Mnih2015} \shortcite{Mnih2015}  and 
van Hasselt et al. \shortcite{van2016deep}. The average scores of the agent are evaluated during training, every 1M steps in the environment, by suspending learning and evaluating the latest agent for 500K frames. Episodes are truncated at 108K frames (or 30 minutes of simulated play), as in van Hasselt et al. \shortcite{van2016deep}.

Agents' scores are normalized, per game, so that 0\% corresponds to a random agent and 100\% to the average score of a human expert. Normalized scores can be aggregated across all Atari levels to compare the performance of different agents. It is common to track the \emph{median} human normalized performance across all games. We also consider the number of games where the agent's performance is above some fraction of human performance, to disentangle where improvements in the median come from. The \emph{mean} human normalized performance is potentially less informative, as it is dominated by a few games (e.g., Atlantis) where agents achieve scores orders of magnitude higher than humans do. 

Besides tracking the median performance as a function of environment steps, at the end of training we re-evaluate the best agent snapshot using two different testing regimes. In the \textit{no-ops starts} regime, we insert a random number (up to 30) of no-op actions at the beginning of each episode (as we do also in training). In the \textit{human starts} regime, episodes are initialized with points randomly sampled from  the initial portion of human expert trajectories \cite{Nair2015}; the difference between the two regimes indicates the extent to which the agent has over-fit to its own trajectories. 

Due to space constraints, we focus on aggregate results across games. However, in the appendix we provide full learning curves for all games and all agents, as well as detailed comparison tables of raw and normalized scores, in both the no-op and human starts testing regimes.

\begin{table}[b]
\centering
\vspace{-0.8em}
\small{
\begin{tabular}{ l | r}
 Parameter        & Value \\ 
\hline
 Min history to start learning  & 80K frames \\
 Adam learning rate       & $0.0000625$ \\ 
 Exploration $\epsilon$   & $0.0$ \\
 Noisy Nets $\sigma_0$ & 0.5 \\
 Target Network Period  &  32K frames \\
 Adam $\epsilon$ & $1.5\times10^{-4}$  \\
 Prioritization type & proportional \\
 Prioritization exponent $\omega$ & $0.5$ \\
 Prioritization importance sampling $\beta$ & $0.4 \rightarrow 1.0$  \\
 Multi-step returns $n$ & 3 \\
 Distributional atoms & $51$ \\
 Distributional min/max values & $[-10, 10]$ \\
\end{tabular}}
\vspace{-0.3em}
\caption{Rainbow hyper-parameters}
\label{table-hyper}
\end{table}

\paragraph{Hyper-parameter tuning.}
All Rainbow's components have a number of hyper-parameters. The combinatorial space of hyper-parameters is too large for an exhaustive search, therefore we have performed limited tuning. For each component, we started with the values used in the paper that introduced this component, and tuned the most sensitive among hyper-parameters by manual coordinate descent.

DQN and its variants do not perform learning updates during the first $200K$ frames, to ensure sufficiently uncorrelated updates. We have found that, with prioritized replay, it is possible to start learning sooner, after only $80K$ frames. 

DQN starts with an exploration $\epsilon$ of 1, corresponding to acting uniformly at random; it anneals the amount of exploration over the first 4M frames, to a final value of 0.1 (lowered to 0.01 in later variants). Whenever using Noisy Nets, we acted fully greedily ($\epsilon=0$), with a value of $0.5$ for the $\sigma_0$ hyper-parameter used to initialize the weights in the noisy stream\footnote{The noise was generated on the GPU. Tensorflow noise generation can be unreliable on GPU. If generating the noise on the CPU, lowering $\sigma_0$ to 0.1 may be helpful.}.  For agents without Noisy Nets, we used $\epsilon$-greedy but decreased the exploration rate faster than was previously used, annealing $\epsilon$ to 0.01 in the first $250K$ frames.

We used the Adam optimizer \cite{kingma2014adam}, which we found less sensitive to the choice of the learning rate than RMSProp. DQN uses a learning rate of $\alpha = 0.00025$ In all Rainbow's variants we used a learning rate of $\alpha/4$, selected among $\{\alpha/2, \alpha/4, \alpha/6\}$, and a value of $1.5\times10^{-4}$ for Adam's $\epsilon$ hyper-parameter.

For replay prioritization we used the recommended proportional variant, with priority exponent $\omega$ of $0.5$, and linearly increased the importance sampling exponent $\beta$ from 0.4 to 1 over the course of training. The priority exponent $\omega$ was tuned comparing values of $\{0.4, 0.5, 0.7\}$. Using the KL loss of distributional DQN as priority, we have observed that performance is very robust to the choice of $\omega$.

The value of $n$ in multi-step learning is a sensitive hyper-parameter of Rainbow. We compared values of $n=1,\ 3,\ \text{and } 5$.  We observed that both $n=3$ and $5$ did well initially, but overall $n=3$ performed the best by the end. 

The hyper-parameters (see Table~\ref{table-hyper}) are identical across all 57 games, i.e., the Rainbow agent really is a \emph{single} agent setup that performs well across all the games.

\section{Analysis}
In this section we analyse the main experimental results. First, we show that Rainbow compares favorably to several published agents. Then we perform ablation studies, comparing several variants of the agent, each corresponding to removing a single component from Rainbow.

\paragraph{Comparison to published baselines.}
In Figure \ref{fig:frontpage} we compare the Rainbow's performance (measured in terms of the median human normalized score across games) to the corresponding curves for A3C, DQN, DDQN, Prioritized DDQN, Dueling DDQN, Distributional DQN, and Noisy DQN. We thank the authors of the Dueling and Prioritized agents for providing the learning curves of these, and report our own re-runs for DQN, A3C, DDQN, Distributional DQN and Noisy DQN. The performance of Rainbow is significantly better than any of the baselines, both in data efficiency, as well as in final performance. Note that we match final performance of DQN after 7M frames, surpass the best final performance of these baselines in 44M frames, and reach substantially improved final performance.

In the final evaluations of the agent, after the end of training, Rainbow achieves a median score of 223\% in the no-ops regime; in the human starts regime we measured a median score of 153\%. In Table \ref{fig:end_of_training_scores} we compare these scores to the published median scores of the individual baselines.

\begin{table}[b!]
\vspace{-0.5em}
\centering
\begin{tabular}{ | l | c c |}
\hline
Agent           & no-ops &  human starts \\
\hline
\hline
 DQN              & 79\%           & 68\% \\
 DDQN (*)            & 117\%          & 110\% \\
 Prioritized DDQN (*)      & 140\%          & 128\% \\
 Dueling DDQN (*)          & 151\%          & 117\% \\
 A3C (*)              &   -     & 116\%     \\
 Noisy DQN          & 118\%           & 102\%   \\
 Distributional DQN   & 164\%          & 125\% \\
 \hline
 Rainbow          & 223\%          & 153\% \\
 \hline

\end{tabular}
\caption{Median normalized scores of the best agent snapshots for Rainbow and baselines.  For methods marked with an asterisk, the scores come from the corresponding publication. DQN's scores comes from the dueling networks paper, since DQN's paper did not report scores for all 57 games. The others scores come from our own implementations. 
}
\label{fig:end_of_training_scores}
\end{table}

In Figure \ref{fig:num_games} (top row) we plot the number of games where an agent has reached some specified level of human normalized performance. From left to right, the subplots show on how many games the different agents have achieved 20\%, 50\%, 100\%, 200\% and 500\% human normalized performance. This allows us to identify where the overall improvements in performance come from. Note that the gap in performance between Rainbow and other agents is apparent at all levels of performance: the Rainbow agent is improving scores on games where the baseline agents were already good, as well as improving in games where baseline agents are still far from human performance.

\paragraph{Learning speed.}
As in the original DQN setup, we ran each agent on a single GPU. The 7M frames required to match DQN's final performance correspond to less than 10 hours of wall-clock time. A full run of 200M frames corresponds to approximately 10 days, and this varies by less than 20\% between all of the discussed variants. The literature contains many alternative training setups that improve performance as a function of wall-clock time by exploiting parallelism, e.g., \citeauthor{Nair2015} \shortcite{Nair2015}, \citeauthor{nes_atari} \shortcite{nes_atari}, and \citeauthor{Mnih:2016} \shortcite{Mnih:2016}. Properly relating the performance across such very different hardware/compute resources is non-trivial, so we focused exclusively on algorithmic variations, allowing apples-to-apples comparisons. While we consider them to be important and complementary, we leave questions of scalability and parallelism to future work.

\begin{figure}[t]
\centering
\vspace*{-.7cm}
\includegraphics[width=\columnwidth, trim={0 0 1cm 0},clip]{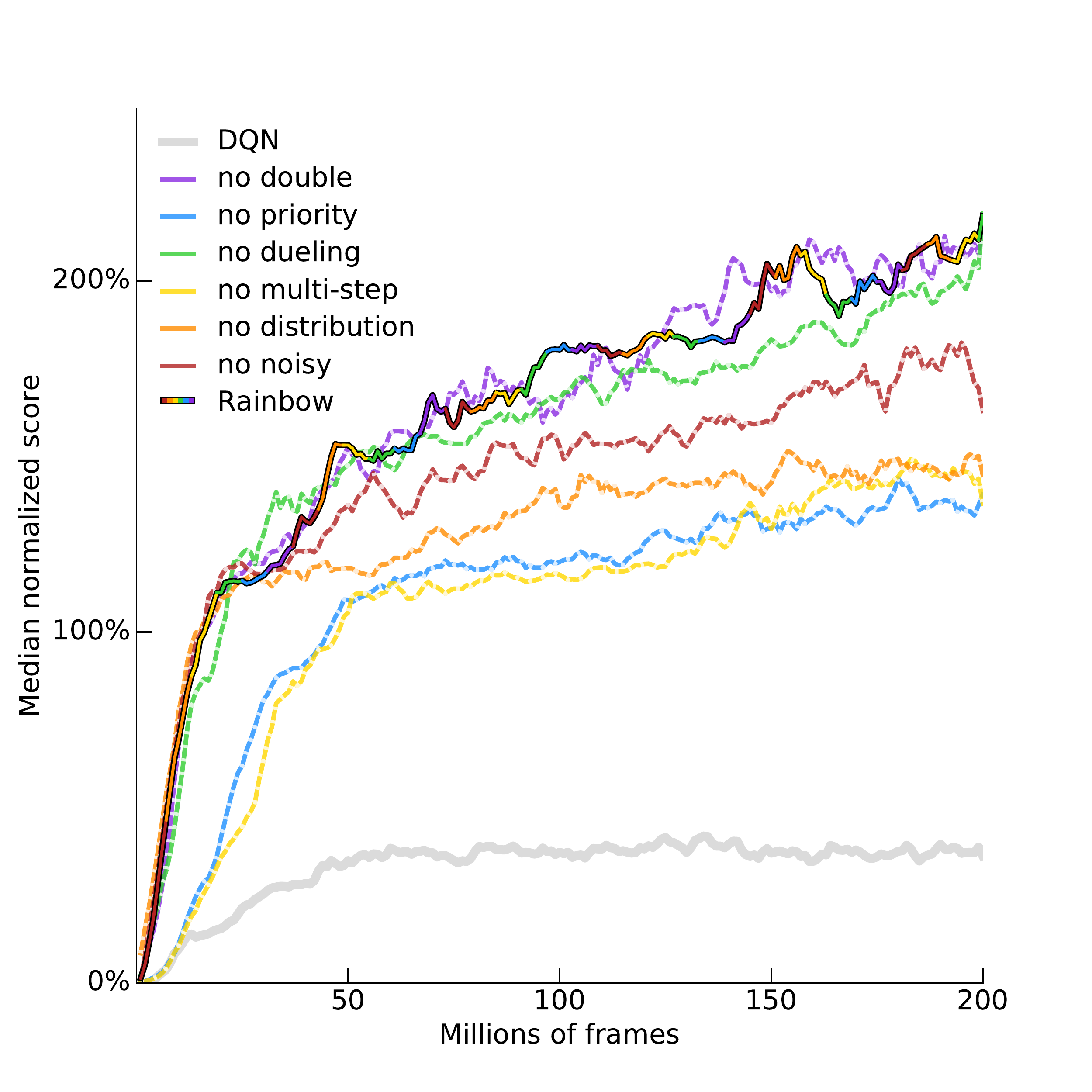}
\vspace*{-0.7cm}
\caption{\textbf{Median human-normalized performance} across 57 Atari games, as a function of time. We compare our integrated agent (rainbow-colored) to DQN (gray) and to  six different ablations (dashed lines). Curves are smoothed with a moving average over 5 points.}
\label{fig:ks_vs_ablations}
\end{figure}

\begin{figure*}[t]
\centering
\includegraphics[width=\textwidth]{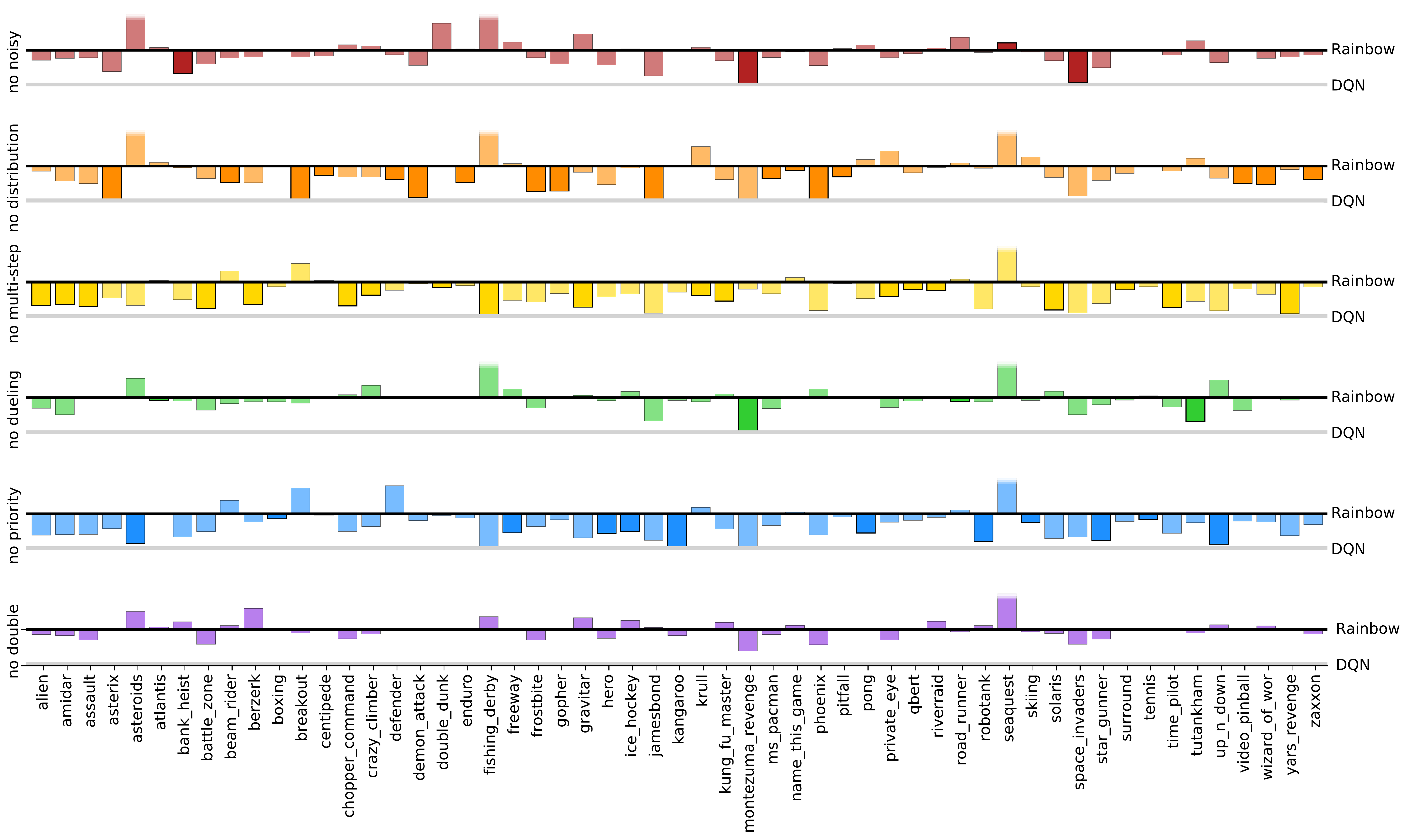}
\caption{\textbf{Performance drops of ablation agents} on all 57 Atari games. Performance is the area under the learning curve, normalized relative to the Rainbow agent and DQN. Two games where DQN outperforms Rainbow are omitted. The ablation leading to the strongest drop is highlighted for each game.
The removal of either prioritization or multi-step learning reduces performance across most games, but the contribution of each component varies substantially per game.}
\label{fig:rainbow_waterfall}
\end{figure*}

\paragraph{Ablation studies.}
\label{ablation}
Since Rainbow integrates several different ideas into a single agent, we conducted additional experiments to understand the contribution of the various components, in the context of this specific combination. 

To gain a better understanding of the contribution of each component to the Rainbow agent, we performed ablation studies. In each ablation, we removed one component from the full Rainbow combination.
Figure~\ref{fig:ks_vs_ablations} shows a comparison for median normalized score of the full Rainbow to six ablated variants. Figure~\ref{fig:num_games} (bottom row) shows a more detailed breakdown of how these ablations perform relative to different thresholds of human normalized performance, and Figure~\ref{fig:rainbow_waterfall} shows the gain or loss from each ablation for every game, averaged over the full learning run.

Prioritized replay and multi-step learning were the two most crucial components of Rainbow, in that removing either component caused a large drop in median performance. Unsurprisingly, the removal of either of these hurt early performance. Perhaps more surprisingly, the removal of multi-step learning also hurt final performance. Zooming in on individual games (Figure~\ref{fig:rainbow_waterfall}), we see both components helped almost uniformly across games (the full Rainbow performed better than either ablation in 53 games out of 57).

Distributional Q-learning ranked immediately below the previous techniques for relevance to the agent's performance. Notably, in early learning no difference is apparent, as shown in Figure~\ref{fig:ks_vs_ablations}, where for the first 40 million frames the distributional-ablation performed as well as the full agent.  However, without distributions, the performance of the agent then started lagging behind.  When the results are separated relatively to human performance in Figure~\ref{fig:num_games}, we see that the distributional-ablation primarily seems to lags on games that are above human level or near it.

In terms of median performance, the agent performed better when Noisy Nets were included; when these are removed and exploration is delegated to the traditional $\epsilon$-greedy mechanism, performance was worse in aggregate (red line in Figure~\ref{fig:ks_vs_ablations}). While the removal of Noisy Nets produced a large drop in performance for several games, it also provided small increases in other games (Figure~\ref{fig:rainbow_waterfall}).

In aggregate, we did not observe a significant difference when removing the dueling network from the full Rainbow. The median score, however, hides the fact that the impact of Dueling differed between games, as shown by Figure~\ref{fig:rainbow_waterfall}. Figure~\ref{fig:num_games} shows that Dueling perhaps provided some improvement on games with above-human performance levels (\# games $> 200\%$), and some degradation on games with sub-human performance (\# games $> 20\%$). 

Also in the case of double Q-learning, the observed difference in median performance (Figure~\ref{fig:ks_vs_ablations}) is limited, with the component sometimes harming or helping depending on the game (Figure~\ref{fig:rainbow_waterfall}). To further investigate the role of double Q-learning, we compared the predictions of our trained agents to the actual discounted returns computed from clipped rewards. Comparing Rainbow to the agent where double Q-learning was ablated, we observed that the actual returns are often higher than $10$ and therefore fall outside the support of the distribution, spanning from $-10$ to $+10$.  This leads to underestimated returns, rather than overestimations.
We hypothesize that clipping the values to this constrained range counteracts the overestimation bias of Q-learning. Note, however, that the importance of double Q-learning may increase if the support of the distributions is expanded.

In the appendix, for each game we show final performance and learning curves for Rainbow, its ablations, and baselines.

\section{Discussion}

We have demonstrated that several improvements to DQN can be successfully integrated into a single learning algorithm that achieves state-of-the-art performance.  Moreover, we have shown that within the integrated algorithm, all but one of the components provided clear performance benefits.
There are many more algorithmic components that we were not able to include, which would be promising candidates for further experiments on integrated agents. Among the many possible candidates, we discuss several below.

We have focused here on value-based methods in the Q-learning family. We have not considered purely policy-based RL algorithms such as trust-region policy optimisation \cite{TRPO}, nor actor-critic methods \cite{Mnih:2016,ODonoghue2016PGQCP}. 

A number of algorithms exploit a sequence of data to achieve improved learning efficiency. Optimality tightening \cite{HeLSP16} uses multi-step returns to construct additional inequality bounds, instead of using them to replace the 1-step targets used in Q-learning. Eligibility traces allow a soft combination over n-step returns \cite{Sutton:1988}. However, sequential methods all leverage more computation per gradient than the multi-step targets used in Rainbow. Furthermore, introducing prioritized sequence replay raises questions of how to store, replay and prioritise sequences. 

Episodic control \cite{Episodic} also focuses on data efficiency, and was shown to be very effective in some domains. It improves early learning by using episodic memory as a complementary learning system, capable of immediately re-enacting successful action sequences.

Besides Noisy Nets, numerous other exploration methods could also be useful algorithmic ingredients: among these Bootstrapped DQN \cite{Osband2016DeepEV}, intrinsic motivation \cite{StadieLA15} and count-based exploration \cite{Bellemare2016UnifyingCE}. Integration of these alternative components is fruitful subject for further research.

In this paper we have focused on the core learning updates, without exploring alternative computational architectures. Asynchronous learning from parallel copies of the environment, as in A3C \cite{Mnih:2016}, Gorila \cite{Nair2015}, or Evolution Strategies \cite{nes_atari}, can be effective in speeding up learning, at least in terms of wall-clock time. Note, however, they can be less data efficient.

Hierarchical RL has also been applied with success to several complex Atari games. Among successful applications of HRL we highlight h-DQN \cite{hDQN} and Feudal Networks \cite{Feudal}.

The state representation could also be made more efficient by exploiting auxiliary tasks such as pixel control or feature control \cite{JaderbergMCSLSK16}, supervised predictions \cite{DosovitskiyK16} or successor features \cite{kulkarni2016deep}.

To evaluate Rainbow fairly against the baselines, we have followed the common domain modifications of clipping rewards, fixed action-repetition, and frame-stacking, but these might be removed by other learning algorithm improvements. Pop-Art normalization \cite{PopNIPS} allows reward clipping to be removed, while preserving a similar level of performance. Fine-grained action repetition \cite{sharma2017learning} enabled to learn how to repeat actions. A recurrent state network \cite{hausknecht2015deep} can learn a temporal state representation, replacing the fixed stack of observation frames. In general, we believe that exposing the real game to the agent is a promising direction for future research.

\bibliographystyle{aaai}
\bibliography{bibs}

\begin{figure*}
\centering
\includegraphics[width=0.9\textwidth]{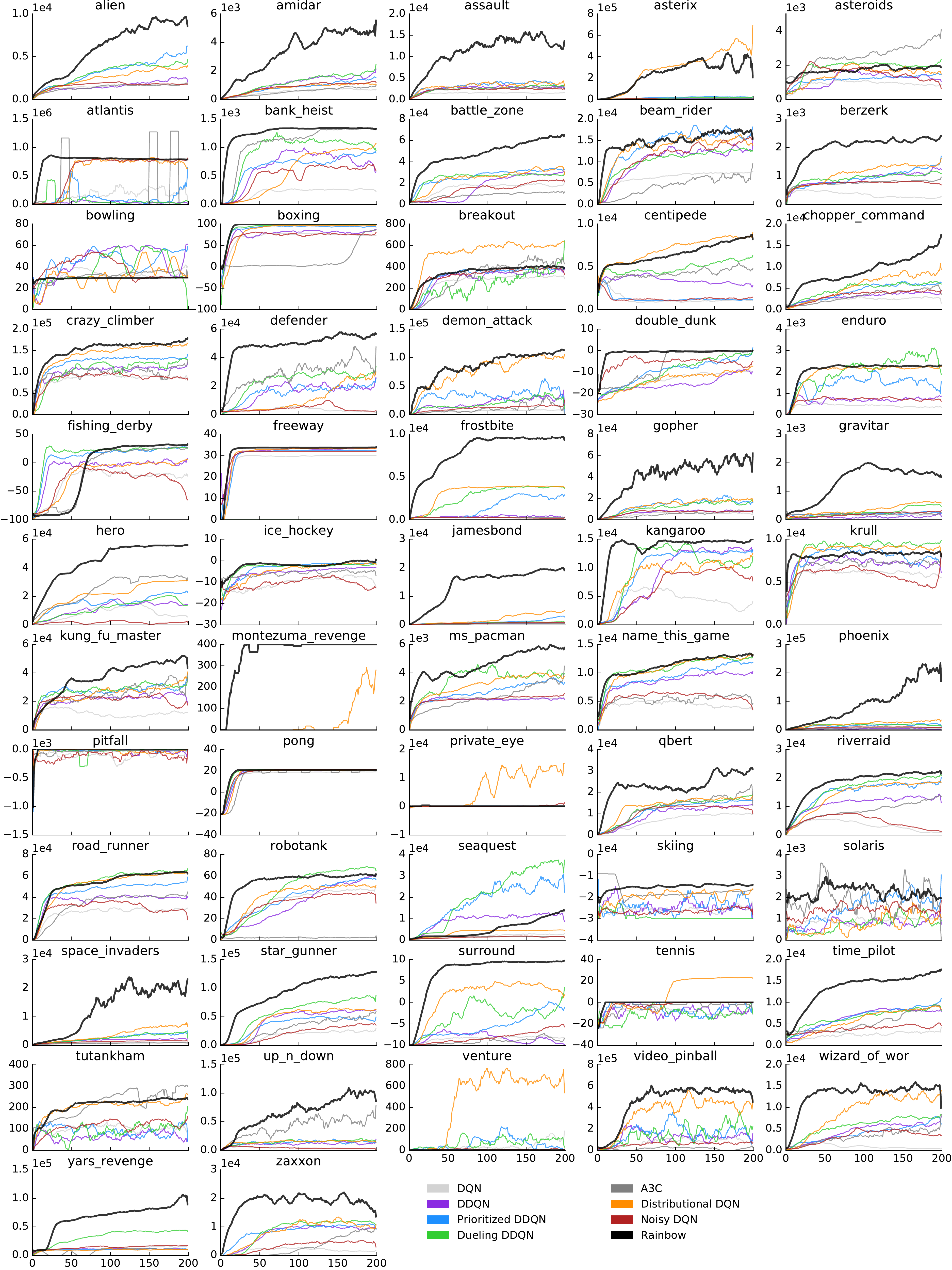}
\caption{Learning curves for Rainbow and the baselines discussed in the paper, for each individual game. Every curve is smoothed with a moving average of 10 to improve readability.}
\label{fig:published_individual}
\end{figure*}

\begin{figure*}
\centering
\includegraphics[width=0.9\textwidth]{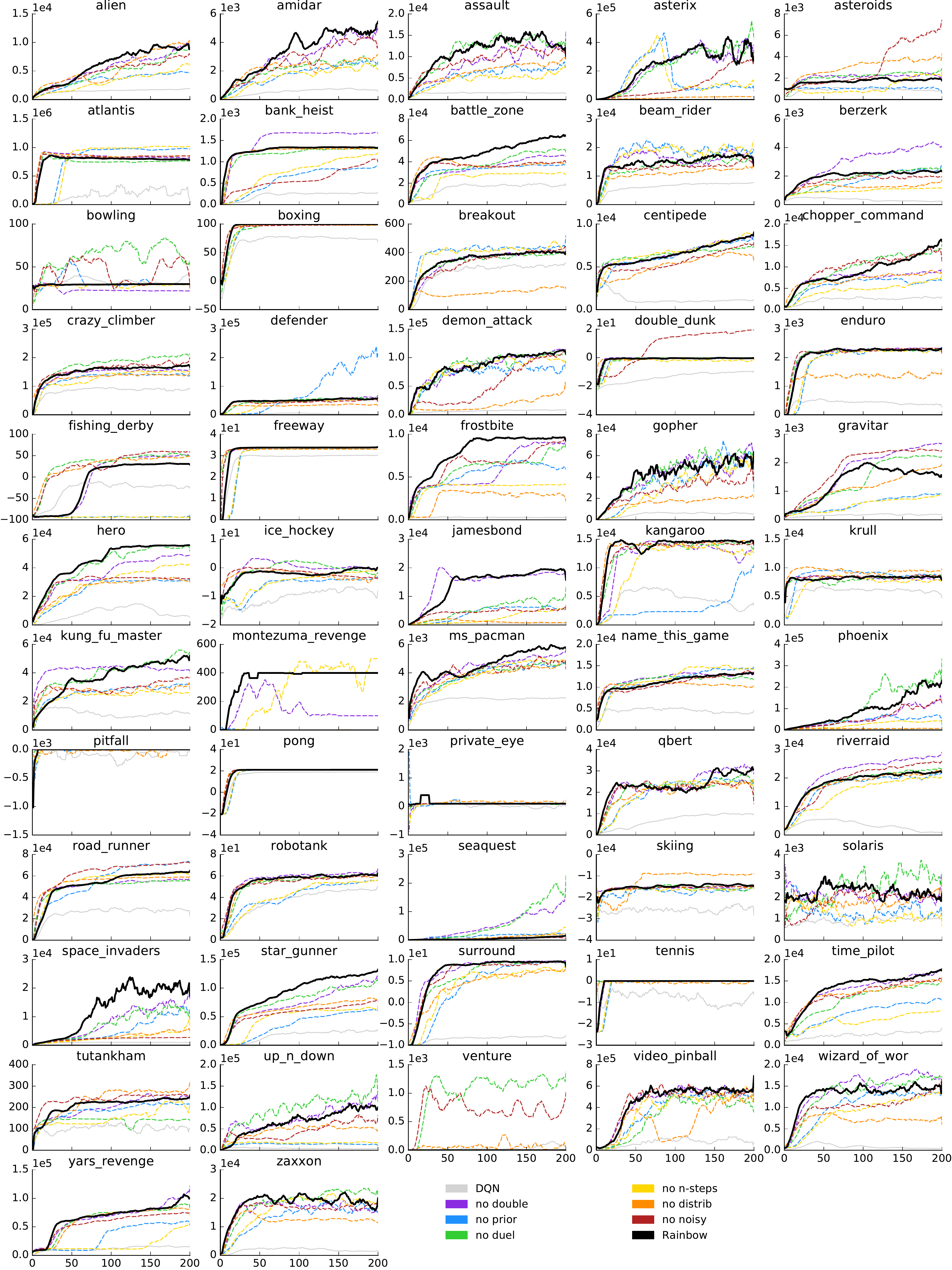}
\caption{Learning curves for Rainbow and its ablations, for each individual game. Every curve is smoothed with a moving average of 10 to improve readability.}
\label{fig:ablations_individual}
\end{figure*}

\onecolumn
\begin{center}
{\Huge{Appendix}}\\
\end{center}
\vspace{1.5cm}
Table \ref{tab:preprocess} lists the preprocessing of environment frames, rewards and discounts introduced by DQN. Table \ref{tab:other_hyperparams} lists the additional hyper-parameters that Rainbow inherits from DQN and the other baselines considered in this paper. The hyper-parameters for which Rainbow uses non standard settings are instead listed in the main text. In the subsequent pages, we list the tables showing, for each game, the score achieved by Rainbow and several baselines in both the no-ops regime (Table \ref{tab:noop}) and the human-starts regime (Table \ref{tab:hs}). In Figures \ref{fig:published_individual} and \ref{fig:ablations_individual} we also plot, for each game, the learning curves of Rainbow, several baselines, and all ablation experiments. These learning curves are smoothed with a moving average over a window of 10.

\vspace{6em}

\begin{table}[h!]
\centering
\begin{tabular}{ l | c }
\hline
Hyper-parameter           & value \\
\hline
 Grey-scaling           & True \\
 Observation down-sampling      & (84, 84) \\ 
 Frames stacked         &  4 \\
 Action repetitions        &  4 \\
 Reward clipping        & [-1, 1] \\
 Terminal on loss of life & True \\
 Max frames per episode & 108K \\
\hline

\end{tabular}
\caption{\textbf{Preprocessing}: the values of these hyper-parameters are the same used by DQN and it’s variants. They are here listed for completeness. Observations are grey-scaled and rescaled to $84\times84$ pixels. 4 consecutive frames are concatenated as each state's representation. Each action selected by the agent is repeated for 4 times. Rewards are clipped between $-1, +1$. In games where the player has multiple lives, transitions associated to the loss of a life are considered terminal. All episodes are capped after 108K frames.
}
\label{tab:preprocess}
\end{table}

\vspace{3em}

\begin{table}[h!]
\centering
\begin{tabular}{ l | c }
\hline
Hyper-parameter           & value \\
\hline
 \ Q network: channels     &  32, 64, 64 \\
 \ Q network: filter size  &  $8\times8, 4\times4, 3\times3$ \\
 \ Q network: stride       &  4, 2, 1 \\
 \ Q network: hidden units         &  512 \\
 \ Q network: output units         & Number of actions \\
  Discount factor               & 0.99 \\
  Memory size            & 1M transitions \\ 
  Replay period          & every 4 agent steps \\
  Minibatch size         & 32 \\
\hline

\end{tabular}
\caption{\textbf{Additional hyper-parameters}: the values of these hyper-parameters are the same used by DQN and it's variants. The network has 3 convolutional layers: with 32, 64 and 64 channels. The layers use $8\times8$, $4\times4$, $3\times3$ filters with strides of 4, 2, 1, respectively. The value and advantage streams of the dueling architecture have both a hidden layer with 512 units. The output layer of the network has a number of units equal to the number of actions available in the game. We use a discount factor of 0.99, which is set to 0 on terminal transitions. We perform a learning update every 4 agent steps, using mini-batches of 32 transitions.
}
\label{tab:other_hyperparams}
\end{table}

\newpage

\begin{table*}[t]
\centering
\scalebox{0.95}{

\begin{tabular}{l|rrrrrrrr} 
\toprule
              Game &       DQN &       A3C &      DDQN & Prior. DDQN &    Duel. DDQN & Distrib. DQN &    Noisy DQN & Rainbow \\
\midrule
             alien &     634.0 &     518.4 &    1033.4 &       900.5 &    1,486.5 &        1,997.5 &      533.3 &   {\bf 6,022.9} \\
            amidar &     178.4 &    {\bf 263.9} &     169.1 &       218.4 &      172.7 &          237.7 &      148.0 &      202.8 \\
           assault &    3489.3 &    5474.9 &    6060.8 &     7,748.5 &    3,994.8 &        5,101.3 &    5,124.3 &   {\bf 14,491.7} \\
           asterix &    3170.5 &   22140.5 &   16837.0 &    31,907.5 &   15,840.0 &     {\bf 395,599.5 }&    8,277.3 &  280,114.0 \\
         asteroids &    1458.7 &    {\bf 4474.5 }&    1193.2 &     1,654.0 &    2,035.4 &        2,071.7 &    4,078.1 &    2,249.4 \\
          atlantis &  292491.0 &  {\bf 911,091.0 }&  319688.0 &   593,642.0 &  445,360.0 &      289,803.0 &  303,666.5 &  814,684.0 \\
        bank\_heist &     312.7 &     970.1 &     886.0 &       816.8 &    {\bf 1,129.3 }&          835.6 &      955.0 &      826.0 \\
       battle\_zone &   23750.0 &   12950.0 &   24740.0 &    29,100.0 &   31,320.0 &       32,250.0 &   26,985.0 &   {\bf 52,040.0 }\\
        beam\_rider &    9743.2 &   22707.9 &   17417.2 &    {\bf 26,172.7} &   14,591.3 &       15,002.4 &   15,241.5 &   21,768.5 \\
           berzerk &     493.4 &     817.9 &    1011.1 &     1,165.6 &      910.6 &        1,000.0 &      670.8 &   {\bf  1,793.4} \\
           bowling &      56.5 &      35.1 &      69.6 &        65.8 &       65.7 &           76.8 &       {\bf 79.3 }&       39.4 \\
            boxing &      70.3 &      59.8 &      73.5 &        68.6 &      {\bf  77.3} &           62.1 &       66.3 &       54.9 \\
          breakout &     354.5 &    {\bf  681.9 }&     368.9 &       371.6 &      411.6 &          548.7 &      423.3 &      379.5 \\
         centipede &    3973.9 &    3755.8 &    3853.5 &     3,421.9 &    4,881.0 &       {\bf  7,476.9 }&    4,214.4 &    7,160.9 \\
   chopper\_command &    5017.0 &    7021.0 &    3495.0 &     6,604.0 &    3,784.0 &        9,600.5 &    8,778.5 &  {\bf  10,916.0 }\\
     crazy\_climber &   98128.0 &  112646.0 &  113782.0 &   131,086.0 &  124,566.0 &      154,416.5 &   98,576.5 &  {\bf 143,962.0 }\\
          defender &   15917.5 &   56533.0 &   27510.0 &    21,093.5 &   33,996.0 &       32,246.0 &   18,037.5 &  {\bf  47,671.3 }\\
      demon\_attack &   12550.7 & {\bf 113,308.4 }&   69803.4 &    73,185.8 &   56,322.8 &      109,856.6 &   25,207.8 &  109,670.7 \\
       double\_dunk &      -6.0 &      -0.1 &      -0.3 &        {\bf  2.7} &       -0.8 &           -3.7 &       -1.0 &       -0.6 \\
            enduro &     626.7 &     -82.5 &    1216.6 &     1,884.4 &    2,077.4 &       {\bf  2,133.4 }&    1,021.5 &    2,061.1 \\
     fishing\_derby &      -1.6 &      18.8 &       3.2 &         9.2 &       -4.1 &           -4.9 &       -3.7 &      {\bf  22.6 }\\
           freeway &      26.9 &       0.1 &      28.8 &        27.9 &        0.2 &           28.8 &       27.1 &    {\bf    29.1 }\\
         frostbite &     496.1 &     190.5 &    1448.1 &     2,930.2 &    2,332.4 &        2,813.9 &      418.8 &   {\bf  4,141.1} \\
            gopher &    8190.4 &   10022.8 &   15253.0 &    57,783.8 &   20,051.4 &       27,778.3 &   13,131.0 & {\bf   72,595.7 }\\
          gravitar &     298.0 &     303.5 &     200.5 &       218.0 &      297.0 &          422.0 &      250.5 &   {\bf    567.5 }\\
              hero &   14992.9 &   32464.1 &   14892.5 &    20,506.4 &   15,207.9 &       28,554.2 &    2,454.2 &  {\bf  50,496.8 }\\
        ice\_hockey &      -1.6 &      -2.8 &      -2.5 &        -1.0 &       -1.3 &          {\bf  -0.1} &       -2.4 &       -0.7 \\
          kangaroo &    4496.0 &      94.0 &  {\bf  11204.0 }&    10,241.0 &   10,334.0 &        9,555.5 &    7,465.0 &   10,841.0 \\
             krull &    6206.0 &    5560.0 &    6796.1 &     7,406.5 &  {\bf   8,051.6 }&        6,757.8 &    6,833.5 &    6,715.5 \\
    kung\_fu\_master &   20882.0 &   28819.0 &   30207.0 &    31,244.0 &   24,288.0 &     {\bf   33,890.0 }&   27,921.0 &   28,999.8 \\
 montezuma\_revenge &      47.0 &      67.0 &      42.0 &        13.0 &       22.0 &          130.0 &       55.0 &    {\bf   154.0 }\\
         ms\_pacman &    1092.3 &     653.7 &    1241.3 &     1,824.6 &    2,250.6 &        2,064.1 &    1,012.1 &   {\bf  2,570.2 }\\
    name\_this\_game &    6738.8 &   10476.1 &    8960.3 &   {\bf  11,836.1} &   11,185.1 &       11,382.3 &    7,186.4 &   11,686.5 \\
           phoenix &    7484.8 &   52894.1 &   12366.5 &    27,430.1 &   20,410.5 &       31,358.3 &   15,505.0 &  {\bf 103,061.6 }\\
           pitfall &    -113.2 &     -78.5 &    -186.7 &     {\bf   -14.8} &      -46.9 &         -342.8 &     -154.4 &      -37.6 \\
              pong &      18.0 &       5.6 &     {\bf  19.1} &        18.9 &       18.8 &           18.9 &       18.0 &       19.0 \\
       private\_eye &     207.9 &     206.9 &    -575.5 &       179.0 &      292.6 &        5,717.5 &   {\bf  5,955.4} &    1,704.4 \\
             qbert &    9271.5 &   15148.8 &   11020.8 &    11,277.0 &   14,175.8 &       15,035.9 &    9,176.6 &  {\bf  18,397.6 }\\
       road\_runner &   35215.0 &   34216.0 &   43156.0 &    56,990.0 &  {\bf  58,549.0 }&       56,086.0 &   35,376.5 &   54,261.0 \\
          robotank &      58.7 &      32.8 &      59.1 &        55.4 &      {\bf  62.0 }&           49.8 &       50.9 &       55.2 \\
          seaquest &    4216.7 &    2355.4 &   14498.0 &   {\bf  39,096.7} &   37,361.6 &        3,275.4 &    2,353.1 &   19,176.0 \\
            skiing &  -12142.1 &  -10911.1 &  -11490.4 &  {\bf  -10,852.8} &  -11,928.0 &      -13,247.7 &  -13,905.9 &  -11,685.8 \\
           solaris &    1295.4 &    1956.0 &     810.0 &     2,238.2 &    1,768.4 &        2,530.2 &    2,608.2 &    {\bf 2,860.7} \\
    space\_invaders &    1293.8 &   {\bf 15,730.5 }&    2628.7 &     9,063.0 &    5,993.1 &        6,368.6 &    1,697.2 &   12,629.0 \\
       star\_gunner &   52970.0 & {\bf  138218.0 }&   58365.0 &    51,959.0 &   90,804.0 &       67,054.5 &   31,864.5 &  123,853.0 \\
          surround &      -6.0 &      -9.7 &       1.9 &        -0.9 &        4.0 &            4.5 &       -3.1 &      {\bf   7.0} \\
            tennis &      11.1 &      -6.3 &      -7.8 &        -2.0 &        4.4 &          {\bf  22.6 }&       -2.1 &       -2.2 \\
        time\_pilot &    4786.0 &  {\bf  12,679.0} &    6608.0 &     7,448.0 &    6,601.0 &        7,684.5 &    5,311.0 &   11,190.5 \\
         tutankham &      45.6 &    {\bf  156.3 }&      92.2 &        33.6 &       48.0 &          124.3 &      123.3 &      126.9 \\
           venture &     136.0 &      23.0 &      21.0 &       244.0 &      200.0 &          {\bf 462.0} &       10.5 &       45.0 \\
     video\_pinball &  154414.1 &  331628.1 &  367823.7 &   374,886.9 &  110,976.2 &      455,052.7 &  241,851.7 & {\bf  506,817.2 }\\
     wizard\_of\_wor &    1609.0 &   {\bf 17,244.0 } &    6201.0 &     7,451.0 &    7,054.0 &       11,824.5 &    4,796.5 &   14,631.5 \\
      yars\_revenge &    4577.5 &    7157.5 &    6270.6 &     5,965.1 &   25,976.5 &        8,267.7 &    5,487.3 &   {\bf 93,007.9 }\\
            zaxxon &    4412.0 &  {\bf  24,622.0} &    8593.0 &     9,501.0 &   10,164.0 &       15,130.0 &    7,650.5 &   19,658.0 \\
\bottomrule
\end{tabular}
}
\caption{\textbf{Human Starts} evaluation regime: Raw scores across all games, averaged over 200 testing episodes, from the agent snapshot that obtained the highest score during training. We report the published scores for DQN, A3C, DDQN, Dueling DDQN, and Prioritized DDQN. For Distributional DQN and Rainbow we report our own evaluations of the agents.}
\label{tab:hs}
\end{table*}

\begin{table*}[t]
\centering
\scalebox{0.95}{

\begin{tabular}{l|rrrrrrr}
\toprule
              Game &       DQN &    DDQN & Prior. DDQN &    Duel. DDQN & Distrib. DQN &  Noisy DQN &  Rainbow \\
\midrule
             alien &    1620.0 &    3747.7 &     6,648.6 &    4,461.4 &        4,055.8 &    2,394.9 &    {\bf 9,491.7} \\
            amidar &     978.0 &    1793.3 &     2,051.8 &    2,354.5 &        1,267.9 &    1,608.0 &   {\bf  5,131.2} \\
           assault &    4280.0 &    5393.2 &     7,965.7 &    4,621.0 &        5,909.0 &    5,198.6 & {\bf   14,198.5} \\
           asterix &    4359.0 &   17356.5 &    41,268.0 &   28,188.0 &      400,529.5 &   12,403.8 & {\bf  428,200.3 }\\
         asteroids &    1364.5 &     734.7 &     1,699.3 &    2,837.7 &        2,354.7 &  {\bf   4,814.1} &    2,712.8 \\
          atlantis &  279987.0 &  106056.0 &   427,658.0 &  382,572.0 &      273,895.0 &  329,010.0 &  {\bf 826,659.5} \\
        bank\_heist &     455.0 &    1030.6 &     1,126.8 &    {\bf 1,611.9 }&        1,056.7 &    1,323.0 &    1,358.0 \\
       battle\_zone &   29900.0 &   31700.0 &    38,130.0 &   37,150.0 &       41,145.0 &   32,050.0 &   {\bf 62,010.0 }\\
        beam\_rider &    8627.5 &   13772.8 &   {\bf  22,430.7 }&   12,164.0 &       13,213.4 &   12,534.0 &   16,850.2 \\
           berzerk &     585.6 &    1225.4 &     1,614.2 &    1,472.6 &        1,421.8 &      837.3 &   {\bf  2,545.6 }\\
           bowling &      50.4 &      68.1 &        62.6 &       65.5 &        {\bf    74.1 }&       77.3 &       30.0 \\
            boxing &      88.0 &      91.6 &        98.8 &       99.4 &           98.1 &       83.3 &      {\bf  99.6} \\
          breakout &     385.5 &     418.5 &       381.5 &      345.3 &        {\bf   612.5} &      459.1 &      417.5 \\
         centipede &    4657.7 &    5409.4 &     5,175.4 &    7,561.4 &      {\bf   9,015.5} &    4,355.8 &    8,167.3 \\
   chopper\_command &    6126.0 &    5809.0 &     5,135.0 &   11,215.0 &       13,136.0 &    9,519.0 &  {\bf  16,654.0 }\\
     crazy\_climber &  110763.0 &  117282.0 &  {\bf  183,137.0} &  143,570.0 &      178,355.0 &  118,768.0 &  168,788.5 \\
          defender &   23633.0 &   35338.5 &    24,162.5 &   42,214.0 &       37,896.8 &   23,083.0 &  {\bf  55,105.0} \\
      demon\_attack &   12149.4 &   58044.2 &    70,171.8 &   60,813.3 &      110,626.5 &   24,950.1 & {\bf  111,185.2} \\
       double\_dunk &      -6.6 &      -5.5 &       {\bf   4.8} &        0.1 &           -3.8 &       -1.8 &       -0.3 \\
            enduro &     729.0 &    1211.8 &     2,155.0 &    2,258.2 &      {\bf   2,259.3 }&    1,129.2 &    2,125.9 \\
     fishing\_derby &      -4.9 &      15.5 &        30.2 &      {\bf  46.4} &            9.1 &        7.7 &       31.3 \\
           freeway &      30.8 &      33.3 &        32.9 &        0.0 &           33.6 &       32.0 &       {\bf 34.0} \\
         frostbite &     797.4 &    1683.3 &     3,421.6 &    4,672.8 &        3,938.2 &      583.6 &   {\bf  9,590.5 }\\
            gopher &    8777.4 &   14840.8 &    49,097.4 &   15,718.4 &       28,841.0 &   15,107.9 &  {\bf  70,354.6} \\
          gravitar &     473.0 &     412.0 &       330.5 &      588.0 &          681.0 &      443.5 &   {\bf  1,419.3} \\
              hero &   20437.8 &   20130.2 &    27,153.9 &   20,818.2 &       33,860.9 &    5,053.1 &  {\bf  55,887.4} \\
        ice\_hockey &      -1.9 &      -2.7 &         0.3 &        0.5 &           {\bf  1.3 }&       -2.1 &        1.1 \\
          kangaroo &    7259.0 &   12992.0 &    14,492.0 &   {\bf 14,854.0 }&       12,909.0 &   12,117.0 &   14,637.5 \\
             krull &    8422.3 &    7920.5 &    10,263.1 &   {\bf 11,451.9} &        9,885.9 &    9,061.9 &    8,741.5 \\
    kung\_fu\_master &   26059.0 &   29710.0 &    43,470.0 &   34,294.0 &       43,009.0 &   34,099.0 &  {\bf  52,181.0 }\\
 montezuma\_revenge &       0.0 &       0.0 &         0.0 &        0.0 &          367.0 &        0.0 &     {\bf  384.0 }\\
         ms\_pacman &    3085.6 &    2711.4 &     4,751.2 &   {\bf  6,283.5} &        3,769.2 &    2,501.6 &    5,380.4 \\
    name\_this\_game &    8207.8 &   10616.0 &   {\bf  13,439.4} &   11,971.1 &       12,983.6 &    8,332.4 &   13,136.0 \\
           phoenix &    8485.2 &   12252.5 &    32,808.3 &   23,092.2 &       34,775.0 &   16,974.3 &  {\bf 108,528.6 }\\
           pitfall &    -286.1 &     -29.9 &         0.0 &        0.0 &           -2.1 &      -18.2 &        0.0 \\
              pong &      19.5 &      20.9 &        20.7 &       {\bf 21.0 }&           20.8 &       {\bf 21.0} &       20.9 \\
       private\_eye &     146.7 &     129.7 &       200.0 &      103.0 &       {\bf 15,172.9 }&    3,966.0 &    4,234.0 \\
             qbert &   13117.3 &   15088.5 &    18,802.8 &   19,220.3 &       16,956.0 &   15,276.3 &   {\bf 33,817.5 }\\
       road\_runner &   39544.0 &   44127.0 &    62,785.0 &  {\bf  69,524.0 }&       63,366.0 &   41,681.0 &   62,041.0 \\
          robotank &      63.9 &      65.1 &        58.6 &      {\bf  65.3 }&           54.2 &       53.5 &       61.4 \\
          seaquest &    5860.6 &   16452.7 &    44,417.4 &  {\bf  50,254.2 }&        4,754.4 &    2,495.4 &   15,898.9 \\
            skiing &  -13062.3 &   -9021.8 &    -9,900.5 &  {\bf  -8,857.4} &      -14,959.8 &  -16,307.3 &  -12,957.8 \\
           solaris &    3482.8 &    3067.8 &     1,710.8 &    2,250.8 &      {\bf   5,643.1} &    3,204.5 &    3,560.3 \\
    space\_invaders &    1692.3 &    2525.5 &     7,696.9 &    6,427.3 &        6,869.1 &    2,145.5 &  {\bf  18,789.0 }\\
       star\_gunner &   54282.0 &   60142.0 &    56,641.0 &   89,238.0 &       69,306.5 &   34,504.5 & {\bf  127,029.0} \\
          surround &      -5.6 &      -2.9 &         2.1 &        4.4 &            6.2 &       -3.3 &      {\bf   9.7} \\
            tennis &      12.2 &     -22.8 &         0.0 &        5.1 &          {\bf  23.6 }&        0.0 &       -0.0 \\
        time\_pilot &    4870.0 &    8339.0 &    11,448.0 &   11,666.0 &        7,875.0 &    6,157.0 & {\bf   12,926.0 }\\
         tutankham &      68.1 &     218.4 &        87.2 &      211.4 &         {\bf  249.4 }&      231.6 &      241.0 \\
           venture &     163.0 &      98.0 &       863.0 &      497.0 &    {\bf    1,107.0} &        0.0 &        5.5 \\
     video\_pinball &  196760.4 &  309941.9 &   406,420.4 &   98,209.5 &      478,646.7 &  270,444.6 &  {\bf 533,936.5}\\
     wizard\_of\_wor &    2704.0 &    7492.0 &    10,373.0 &    7,855.0 &       15,994.5 &    5,432.0 &   {\bf 17,862.5 }\\
      yars\_revenge &   18089.9 &   11712.6 &    16,451.7 &   49,622.1 &       16,608.6 &    9,570.1 &  {\bf 102,557.0 }\\
            zaxxon &    5363.0 &   10163.0 &    13,490.0 &   12,944.0 &       18,347.5 &    9,390.0 &  {\bf  22,209.5 }\\
\bottomrule
\end{tabular}
}
\caption{\textbf{No-op starts} evaluation regime: Raw scores across all games, averaged over 200 testing episodes, from the agent snapshot that obtained the highest score during training. We report the published scores for DQN, DDQN, Dueling DDQN, and Prioritized DDQN. For Distributional DQN and Rainbow we report our own evaluations of the agents. A3C is not listed since the paper did not report the scores for the no-ops regime.}
\label{tab:noop}
\end{table*}

\end{document}